\documentclass[final,3p,times,twocolumn]{elsarticle}
\usepackage{amssymb}
\usepackage{tabularx}
\usepackage{multirow}
\usepackage{diagbox}
\usepackage{stfloats}
\usepackage[dvipsnames]{xcolor}
\usepackage{graphicx} %package to manage images
\graphicspath{ {image/} }
\journal{CAD}

\begin{document}
\begin{frontmatter}



\title{Data-driven Upsampling of Point Clouds}


\author{Wentai Zhang}
\author{Haoliang Jiang\fnref{cor2}}
\author{Zhangsihao Yang\fnref{cor2}}
\author{Soji Yamakawa}
\author{Kenji Shimada}
\author{Levent Burak Kara\corref{cor1}}

\fntext[cor2]{Equal contribution.}
\cortext[cor1]{Address all correspondences to \ead{lkara@cmu.edu}}

\address{Department of Mechanical Engineering, Carnegie Mellon University, Pittsburgh, PA, 15213, USA}
%%%%%%%%%%%%%%%%%%%%%%%%%%%%%%%%%%%%%%%%%%%%%%%%%%%%%%%%%%%%%%
\begin{abstract}
High quality upsampling of sparse 3D point clouds is critically useful for a wide range of geometric operations such as reconstruction, rendering, meshing, and analysis. In this paper, we propose a data-driven algorithm that enables an upsampling of 3D point clouds without the need for hard-coded rules. Our approach uses a deep network with Chamfer distance as the loss function, capable of learning the latent features in point clouds belonging to different object categories. We evaluate our algorithm across different amplification factors, with upsampling learned and performed on objects belonging to the same category as well as different categories. We also explore the desirable characteristics of input point clouds as a function of the distribution of the point samples. Finally, we demonstrate the performance of our algorithm in single-category training versus multi-category training scenarios. The final proposed model is compared against a baseline, optimization-based upsampling method. Results indicate that our algorithm is capable of generating \textcolor{black}{more accurate upsamplings with less Chamfer loss}.
\end{abstract}

\begin{keyword}


%% or \MSC[2008] code \sep code  (2000 is the default)
point cloud, upsampling, deep learning, neural network
\end{keyword}
%%%%%%%%%%%%%%%%%%%%%%%%%%%%%%%%%%%%%%%%%%%%%%%%%%%%%%%%%%%%%%%
\end{frontmatter}


\section{Introduction}
% Upsampling is a technique to provide more details and more visual quality, such as increasing the number of points of a point cloud, enhancing the resolution of an image, and improving the density of a mesh. Point cloud is widely available due to the boost of 3D sensors and is prevalent in virtual reality, autonomous driving, address mapping and relic preservation. So point cloud upsampling has become an important issue nowadays.

With the emergence of 3D depth sensing technology, point cloud capture has become increasingly common in many applications involving shape digitization and reconstruction. While point cloud quality and point density have a critical impact on the subsequent digital design and processing steps, the large variety of sensing technologies coupled with varying characteristics of the object surfaces and the environment makes high-quality point cloud capture a difficult task. As such, to aid in reconstruction, digitally upsampling an input point cloud to produce a denser representation that remains true to the underlying object is a very desirable capability. However, it remains difficult to do so due to the need to \textit{add} information that does not exist in the input.

%Works to date have explored  learning a latent representation from the input point clouds to for tasks such as object classification and segmentation, while few works focus on how to improve the quality of point cloud in a learning manner. 

% dominant?
%Dense point cloud can provide more detail information when used in analysis. The issue of point cloud upsampling arouses substantial research interest, but it is still challenging to accommodate the application requirements.

Given a point cloud, two common approaches to reconstruction involve direct triangulation, and patch or field regression. Amenta et al. \cite{Amenta:1998:NVS:280814.280947} introduce a reconstruction method based on the three-dimensional Voronoi diagram and Delaunay triangulation. In their algorithm, a set of triangles is generated based on the sample points. Other approaches use interpolating surfaces. Alexa et al. \cite{Alexa:2003:CRP:614289.614541} demonstrate the idea of computing the Voronoi diagram on the moving least squares (MLS) surface and adding new points at vertices of this diagram. Likewise, implicit models have also been extensively used. Apart from classical Poisson, Wavelet and radial basis functions \cite{Berger:2013:BSR:2451236.2451246}, complex fitting strategies such as edge-aware point set resampling (EAR) \cite{Huang:2013:EPS:2421636.2421645} have been explored. 

Such methods are effective at reconstruction when the point clouds are sufficiently dense. However, if the point clouds are so sparse that key shape structures are missing or incomplete, these methods are not likely to recover the missing details as the smoothness between the sample points is usually assumed for cost function minimization or regularization. 

Recently, data-driven methods have been used toward reconstruction from point clouds. Remil et al. \cite{DBLP:journals/corr/RemilXXXW17} present a new approach that learns exemplar priors from model patches. Then these nearest neighbor shape priors from the learned  library can be acquired for each local subset of a given point set. After an appropriate deformation and assembly of the chosen priors, models from the same category as the priors can be reconstructed. Yu et al. \cite{DBLP:journals/corr/abs-1801-06761} develop a neural network called  PU-Net to upsample an input point cloud. The network learns point patches extracted/cropped from the point clouds. A joint loss function is utilized in the training process which constrains the upsampled points to be located on the objective surface and distributed uniformly. To the best of our knowledge, PU-Net is the only current data-driven approach to point cloud upsampling. However, the patch-based learning algorithm of PU-Net places high demands on the resolution of the initial point cloud. 
% Also, they used constant upsampling ratio in all their experiments.

In this work, we aim to learn an upsampling strategy using the point clouds of entire objects rather than patches of individual objects. Specifically, we explore how the information contained in the objects belonging to the same categories impact the upsampling success. As an example of operating on the point clouds of full objects,  Qi et al. \cite{DBLP:journals/corr/QiSMG16} developed a deep learning architecture called PointNet that learns features of point clouds tailored for classification and segmentation. Later, they introduce a hierarchical feature learning neural network named PointNet++ \cite{DBLP:journals/corr/QiYSG17}  capable of extracting both global and local geometric features, with very compelling results.

Achlioptas et al. \cite{DBLP:journals/corr/AchlioptasDMG17} propose a generative point net model based on PointNet. This generative model is designed to capture the latent generative features of the point clouds used for training using an encoder-decoder architecture. In their recent work, they mention shape completion as one of the potential applications for their network. Nevertheless, there is no further exploration in upsampling conditions and the categories of objects. 

In this work, we build on and extend Achlioptas et al.'s work to deploy an upsampling method designed for different object categories and different upsampling amplification factors (AF).  Furthermore, we study the attributes of the input clouds that lead to the most accurate upsampled point clouds. Finally, we expand the encoded input point information to incorporate the vertex normals obtained from the original mesh files and evaluate its influence on the reconstruction performance. In our experiments, models from seven categories in ShapeNetCore \cite{DBLP:journals/corr/ChangFGHHLSSSSX15} are utilized in the training and testing process. The results reveal that data-driven upsampling of sparse point clouds can indeed benefit significantly from categorical class information and moreover, the richness in the data (as obtained through multi-class training) results in high-quality upsampled models for a variety of object categories. 
%(xxx this whole paragraph not sound too strong. I am left with thinkibg that we are not doing anything new network structure wise. We are just taking an existing structure and doing some parametric studies. is this so?).

% Noted that, for PointNet and models based on it, the size of the input of neural network is the same or larger than the size of the output, no experiments on upsampling is reported in these works.
% Based on  (),we modify the network architect to generate a dense pc given a sparse input pc. In this study, the output ground truth is a dense pc uniformly sampled from the original meshes in the dataset. Input pc is a small subset of the groud truth with uniform sampling or curvature-based sampling. The effect of different sampling methods on the upsampling performance are demonstrated. Further, we combine two sampling methods above with various mixing ratios and find an optimal sampling strategy based on the test reconstruction loss. To test the consist of our algorithm, 3 different upsampling ratios are  in the evaluation. 
%   This global level learning method shows more outstanding abilities in dealing with higher upsampling ratios and completion of common shapes.

The key contributions of our work are as follows:
\begin{itemize}
    \item We propose a deep learning algorithm for learning point cloud upsampling using entire object models (rather than patches) as input;
    \item We demonstrate the effect of input point distribution on upsampling quality;
    \item We demonstrate the performance of our approach with diverse amplification factors and the flexibility of our algorithm with single and multiple category training scenarios.
\end{itemize}

% Data-driven method is seldom used in studies of point cloud. Connection of points in a point cloud is relatively weak. Also, they do not contain any grid structure. These properties of point clouds block many explorations in using data-driven methods like deep neural networks to process point clouds. Nevertheless, Qi et al. came up with a deep neural network called PointNet to learn features of point clouds for classification and segmentation.  ()Later, they introduced a hierarchical feature learning neural network named as PointNet++ which is capable of extracting both global and local geometry features.  ()This architecture is proved to be the most outstanding technique for segmentation and classification of point clouds. Noted that, for PointNet and PointNet++, the size of the input of nueral network is the same as the size of the output, no experiments on upsampling was reported in their work.

%One of the drawbacks in similar works is that the triangular model usually requires very thick point sets but the results are likely to be rough and containing some undesirable features like holes and tunnels.
%%%%%%%%%%%%%%%%%%%%%%%%%%%%%%%%%%%%%%%%%%%%%%%%%%%%%%%%%%%%%%%%%%%%%%
\section{Technical Approach}
\subsection{Network Architecture}
The neural network, depicted in Figure \ref{tab: inner}, produces a dense point cloud by taking as input a sparse point cloud of an object. The input is an $N\times M$ matrix where N is the number of input points, and M is the input dimension of one point, which is either 3 or 6 with normal vectors. The encoder is  composed of 1-Dimensional (1D) convolutional layers with filter size 1, followed by a batch normalization \cite{DBLP:journals/corr/IoffeS15} layer and a ReLU \cite{Hinton_rectifiedlinear}. In each layer, the weights and bias of the convolutions are shared among all the points. In the last layer of the encoder, maxpooling is applied along the channel dimension to produce a latent feature vector. Then the feature vector is passed through three fully connected (fc) layers with two ReLU layers in between to complete the reconstruction.

\begin{figure}[!tp]
    \includegraphics[width=8cm,height=2.4cm]{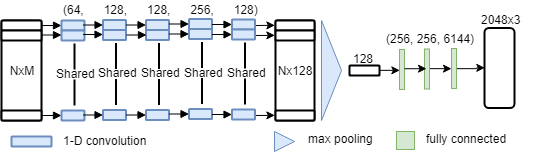}
    \caption{\label{tab: archi}\textcolor{black}{Network Architecture.} The output dimension of each layer is displayed above layer blocks. The five blue blocks represent five 1-D convolutional layers with the output dimension of 64, 128, 128, 256, 128. Each 1-D convolutional layer is followed by ReLU and batch-normalization layers. The three green blocks represents three fully connected layers. A ReLU is placed between two fully connected layers. Reshaping is employed in the last layer. }
% The network employs 1-D convolution followed by ReLU and Batchnormaliazion layers and max pooling to capture the features from models. Then it expands the feature to a thick point cloud fully connected layers and ReLU layers
    
\end{figure}
\subsection{Loss function}
Chamfer distance (CD) \cite{DBLP:journals/corr/FanSG16} and Earth Mover's distance (EMD) \cite{Rubner:2000:EMD:365875.365881} are two loss functions most commonly used in training deep neural networks pertinent to point clouds. 

The Chamfer distance is defined as
\begin{displaymath}
	d_{CD}(S_1,S_2) = \sum\limits_{x\in S_1}\min_{y\in S_2}\|x-y\|^2 + \sum\limits_{y\in S_2}\min_{x\in S_1}\|x - y\|^2,
\end{displaymath}
where subsets $S_1, S_2 \subseteq R^3$ are two point clouds. Chamfer Distance is designed to measure the similarity of two point clouds. The main idea is to find the nearest neighbor for each point in the other point cloud (and vice versa) and sum the squared distances. 

The Earth Mover's distance is defined as
\begin{displaymath}
	d_{EMD}(S_1,S_2) =  \min\limits_{\phi: S_1 \to S_2} 	\sum\limits_{x\in S_1}\|x - \phi(x)\|,
\end{displaymath}
where $\phi$ is a bijection of equal size subsets $S_1, S_2 \subseteq R^3$. There exists a unique and invariant optimal bijection $\phi$ for the point pairs in two sets to move to each other. EMD is a measure of the distance of two point clouds based on the unique bijection $\phi$.

After comparison, we adopt the Chamfer distance as our loss function primarily due to its simplicity and due to its better reconstruction qualities reported by P. Achlioptas et al. \cite{DBLP:journals/corr/AchlioptasDMG17}, who study the performance of shape completion when using CD or EMD as loss functions respectively. Chamfer distance achieves much higher accuracy with little loss in coverage in all evaluated categories of objects. \textcolor{black}{The accuracy is defined as the fraction of predicted points that are within a given radius from any point in the ground truth. The coverage represents the fraction of points in the ground truth which are within the same given radius from any predicted point.} Further details are provided in \ref{app: acc_cov}. 

% hat is (a) Accuracy: which is the fraction of thepredicted points that are within a given radius (ρ) fromany point in the ground truth point cloud and (b) Cov-erage:  which is the fraction of the ground-truth pointsthat are withinρfrom any predicted point. 
%For the categories of objects they evaluated, 

%For the categories of objects they evaluated, the networks trained with CD show better reconstruction qualities. 

%The metric concepts of accuracy and coverage are utilized to assess the completion of the generated point clouds.
% Further details are provided in Appendix \ref{app: acc_cov}(xxx which one?, A, B?).

%For all categories of objects we evaluated, the networks trained with CD show better reconstruction qualities with respect to accuracy and the average of accuracy and coverage 
%(xxx what does average of accuracy and coverage mean? provide a definition of coverage here). 
% The accuracy is defined as the fraction of points in predicted point cloud appear in a radius from points in the ground truth. The coverage was similarly defined but from the ground truth to predicted point cloud. 

\begin{figure}[!tp]
  \centering
  \includegraphics[trim = 0in 0in 0in 0in, clip, width=\columnwidth]{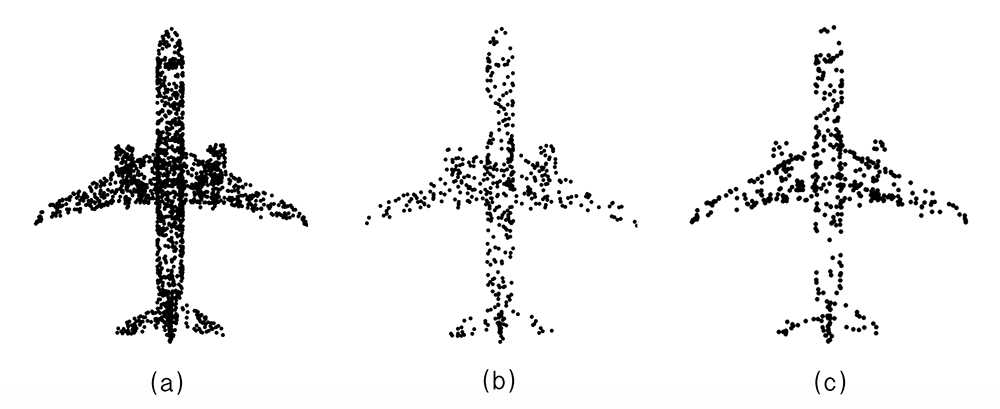}
  \caption{Example results of the two subsampling strategies for the input point clouds.  (a) Ground truth point cloud.  (b) Uniform subsampling.  (c) Curvature-based subsampling.}
  \label{fig:sample}
\end{figure}

%%%%%%%%%%%%%%%%%%%%%%%%%%%%%%%%%%%%%%%%%%%%%%%%%%%%%%%%%%%%%%%%%%%%%%%
\section{Experiments}
\subsection{Dataset and Implementation}

% We use models that are axis aligned and centered with the length of longest side equal to 1. They are from 7 categories of ModelNet40 \cite{DBLP:journals/corr/ChangFGHHLSSSSX15} which are cars, airplanes, boats, benches, chairs, lamps and tables. First, these 7 categories have more than 1000 models each which is benefit for training on single category. Second, the models of these categories ranged from complex to simple and sharp edged (steep) to fairshaped (smooth) could help testing the robustness and flexibility of our algorithm. To extract point clouds from models, two different sampling methods are applied (Section x.x). Unless otherwise specified, we have divided the data of each category with train/validation/test sets of 85\%-5\%-10\% split (Section x.x). For the optimization, we train the network for 2000 epoch using the Adam \cite{DBLP:journals/corr/KingmaB14} algorithm with a learning rate of 0.0005 and a batch size of 50. Generally, the training takes about 40 to 100 minutes on NVIDIA GeForce GTX 1080 Ti GPU based on training size.

%(xxx don't use the word robustness. it is not measurable. people get annyed by that abtract term)
ShapeNetCore is a large-scale 3D CAD Dataset collected and processed by Chang et al. \cite{DBLP:journals/corr/ChangFGHHLSSSSX15}. The dataset consists of 55 shape categories. The orientations of the mesh files are aligned. Also, the models are size-normalized by the longest dimension. We select seven categories from ShapeNetCore as our training data: cars, airplanes, boats, benches, chairs, lamps and tables. These seven categories each have more than 1,000 models. The models from these categories ranged from simple to complex, thereby allowing a performance test of our approach. After selecting a balanced training model set, we divided the data of each category with train/validation/test sets of 85\%-5\%-10\% split. Furthermore, we fix our test cases for all evaluation processes. For learning, we train our neural network for approximately 2,000 epochs using the Adam optimizer \cite{DBLP:journals/corr/KingmaB14} with a learning rate of 0.0005 and a batch size of 50. For each category, the training ranges from 40 to 100 minutes using an NVIDIA GeForce GTX 1080 Ti GPU.

\subsection{Data Pre-processing}
To prepare the point cloud data, for each object, we  uniformly sample a point cloud on the original mesh polygons with 2,048 points (larger polygons getting proportionally more samples). Note that the 2,048 points on these models were selected to strike a balance between upsampled model complexity and computational efficiency for the parametric studies discussed in this study. With the insights arrived at with the study, the network architecture can be altered to change the target number of points in the upsampled models.  In the subsequent experiments, we take these point clouds as our high-resolution ground-truth data. To study the influence of point distributions in the sparse, input point clouds, we downsample each point cloud to 256, 512 and 1,024 points using two approaches: Uniform subsampling (U) or curvature-based subsampling (CB). These subsampled point clouds serve as the input models that our approach aims to upsample. The insight we attain at the end of the study allows a determination of which of the two subsampling approaches (U versus CB) is more suitable for creating accurate upsampled models. 

In Figure \ref{fig:sample}, we show the results of the uniform and curvature-based subsampling respectively.\textcolor{black}{ We follow the sampling methods in [8] and implement two relatively efficient and effective methods to sample the point clouds.}
Monte-Carlo random sample method is used to obtain uniformly distributed points. For curvature sensitive sampling, \textcolor{black}{we first compute a scalar curvature for each edge in the mesh based on \cite{taubin1995}. Then, for each vertex, we take the average of the absolute curvature values of all the edges connecting to it as the its curvature.} When 2,048 points are sampled, each point is assigned a curvature, which is the linear interpolation of curvatures of the three vertices from the same polygons. Then, input points are sampled from a distribution proportional to these curvatures. 
%(xxx more explanation is needed. are you computing the normal curvature or Gaussian curvature? how do you assign curvature to the point sampled on the triangles etc.).

\subsection{Experiment Design}\label{Experiment Details}

\begin{figure*}[!tp]
  \centering
  \includegraphics[trim = 0in 0in 0in 0in, clip, width=\textwidth]{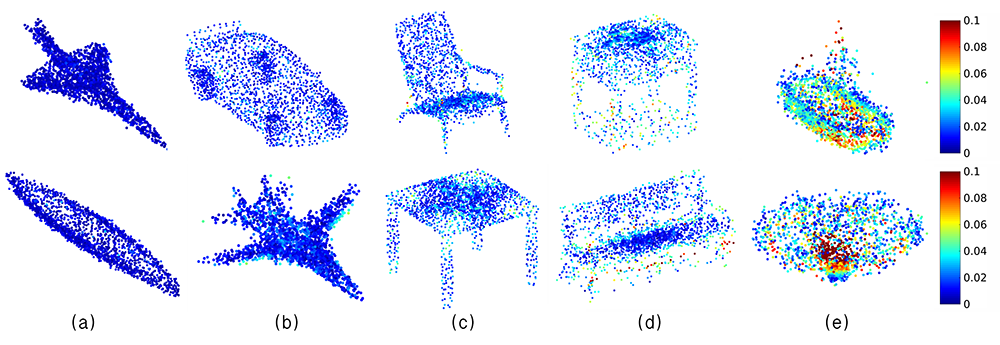}
  \caption{Sample reconstructions from our algorithm with different average test Chamfer loss (ATCL).  (a) ATCL$=0.1\times10^{-3}$.  (b) ATCL$=0.5\times10^{-3}$  (c) ATCL$=1.0\times10^{-3}$  (d) ATCL$=2.0\times10^{-3}$  (e) ATCL$=5.0\times10^{-3}$. The color means the distance between a point in the upsampled point cloud and its nearest point in the ground truth point cloud.}
  \label{fig:reconstruction}
\end{figure*}

\textbf{Single category training and inner-class evaluation}  To demonstrate the effectiveness of our network, we conduct experiments under various upsampling amplification factors including 2, 4 and 8.
%Additionally, we consider to switch point distribution strategies and add additional normal information of points to discover the most competent case (xxx this sentence is unclear). 
For each amplification factor, we input uniformly distributed points and points sampled by curvature-based sampling method respectively. Moreover, we develop another six cases where normal information of the points is taken as additional input. Overall, twelve cases are created for each category where the best case will be picked for further experiments.

\textbf{Single category training and inter-class evaluation}  The results of this experiment demonstrate the ability to perform upsampling in previously unseen categories. Models are trained by only one input category and evaluated on the remaining six categories. We apply the condition under which the network models perform best in the first experiment.
% We do the inter-class experiments using this ratio.
% , where is using uniformly downsampled points without normal information shown in table x, we find that when the upsampling ratio is 8, the results show most obviously the influence of different categories. For this reason, we set the upsampling ratio to 8 in this and next experiments. 

\textbf{Multi-category training and evaluation}   We randomly pick 1,000 models from each category and train our network on the extracted 7,000 point clouds. Then the network is evaluated on the test cases from each category separately. This experiment aims to study the performance of our network when trained with all available object categories but tested on models that were previously unseen (though still belonging to one of the seven object categories). 

%%%%%%%%%%%%%%%%%%%%%%%%%%%%%%%%%%%%%%%%%%%%%%%%%%%%%%%%%%%%%%%%%%%%%%%
\section{Results and Discussions}
\subsection{Inner-class evaluation}\label{inner eval}
% As is stated in
% % section?
% \ref{Experiment Details}, we conduct parametric studies with 12 different cases to test (evaluate?) the upsampling quality (ability?) of our algorithm when trained and tested on point clouds from the same category. The resulting average test Chamfer losses
% % Chamfer loss or distance for the whole paper?
% for each category are shown in table \ref{tab: inner}. The best case for each amplification factor in each category has been marked in bold. 
% % The best case with equal amplification factor of each category 
% Among all the best cases, the largest (biggest) test Chamfer loss value is less than $1.7\times10^{-3}$ (minimum $0.285\times10^{-3}$). When amplification factor is doubled, the most significant increase in Chamfer loss is only around 8\%. Referring to the visualizations of reconstruction quality with various Chamfer loss values, we can deem that overall our upsampling algorithm is excellent and stable in addressing upsampling tasks of different models with multiple amplification factors. 
As stated in section \ref{Experiment Details}, we conduct parametric studies with twelve different cases to evaluate the upsampling performance of our algorithm. The resulting average test Chamfer losses for each category are shown in Table \ref{tab: inner}. The best performances for a given amplification factor of each category is shown in bold. We report the accuracy and coverage values as well in \ref{app: acc_cov}, 
%(xxx created double appendix word in the pdf) 
which follow a pattern similar to that shown in Table \ref{tab: inner}.

Among all the best cases, the largest test Chamfer loss is less than $1.7\times10^{-3}$  (minimum $0.285\times10^{-3}$). When the amplification factor is doubled, the largest increase in Chamfer loss is around 8\%. Based on Figure \ref{fig:reconstruction}, the visualizations of reconstruction quality across different Chamfer loss values, we conclude that overall our upsampling algorithm performs across different objects and multiple amplification factors. Further results are shown in \ref{more results}.

An interesting observation is that input point clouds that are uniform outperform point clouds that have curvature-based samples in all cases where the amplification factor is 4 or 8. But in four of the seven categories, curvature-based sampled point clouds result in better reconstruction quality at $AF=2$. Since Qi at el. \cite{DBLP:journals/corr/QiSMG16}  mention that the points contributing more to the object features usually lie around the edges and outlines of the object, we originally expected points clouds subsampled based on curvature to be more effective at accurate upsampling. 

%(xxx this is a weird paragaraph. the second sentence implies that points near the edges are better, but the whole paragraph is trying to say the opposite. Im confused. Also, edges is not equal to high curvature.) 
% I forgot to put on af=2. This paragraph means we originally expected CB would be better. But we only observe it in 4 out 7 cases when af = 2.

A possible reason is that the input points congregate too much on the edges so that the resulting point cloud is nonuniform. This nonuniformity may decrease the error between the upsampled point cloud and the ground truth since features like sharp edges or high curvature regions require denser point distributions for the features to be faithfully captured. Meanwhile, it inevitably gives rise to the error distance from ground truth to prediction. Because the total number of points is fixed, there must be fewer points in the flat regions. In cases where the amplification factor is small, the side effect of this nonuniformity is not distinctly revealed since the total number of points is still adequate to cover the flat regions. As AF becomes larger, the unbalanced point distribution eventually results in a dominant threat to the upsampling performance. To verify our hypothesis explained above, a hybrid sampling method with various combining ratios is introduced in section \ref{further}. However, as the results demonstrate, normal information does not help to improve the upsampling quality in all the cases. 
%(xxx rewrite)

\begin{table*}[!htb]
\centering
\caption{\label{tab: inner} Evaluation of networks trained on seven categories respectively under twelve conditions. \textcolor{black}{The table shows the evaluation results of single category training and inner-class evaluation described in section 3.3. The lowest test Chamfer loss for each amplification factor is marked in bold. The table indicates that input point clouds that are uniform outperform point clouds that have curvature-based samples in all cases where the amplification factor is 4 or 8. But in four of the seven categories, curvature-based sampled point clouds result in better reconstruction quality at $AF=2$. The corresponding coverage and accuracy results can be found in Table A.5 and Table A.6.}}
\renewcommand{\arraystretch}{2}
\begin{tabular}{ |c||c|c|c|c|c|c|c|c|c|c|c|c|}
\hline
\multirow{3}{*}{} AF & \multicolumn{4}{c|}{2} & \multicolumn{4}{c|}{4} & \multicolumn{4}{c|}{8}\\ 
\cline{1-13}
                    Sample & \multicolumn{2}{c|}{U} & \multicolumn{2}{c|}{CB} & \multicolumn{2}{c|}{U} & \multicolumn{2}{c|}{CB} & \multicolumn{2}{c|}{U} & \multicolumn{2}{c|}{CB}\\
\cline{1-13}
                    Normal & No & Yes & No & Yes & No & Yes & No & Yes & No & Yes & No & Yes\\
\hline
Category & \multicolumn{12}{c|}{Test Chamfer Loss  ($\times10^{-3}$)}\\
\hline
\hline
Airplane & \textbf{0.285} & 0.301 & 0.289 & 0.311 & \textbf{0.294} & 0.316 & 0.313 & 0.343 & \textbf{0.319} & 0.340 & 0.356 & 0.402\\ 
 \hline
Bench & 0.814 &	1.154 &	\textbf{0.771} &	2.480 &	\textbf{0.822} &	1.089 &	0.940 &	1.297 &	\textbf{0.865} &	1.273 &	0.996 &	1.391\\ 
 \hline
Boat & 0.923 &	1.325 &	\textbf{0.899} &	1.410 &	\textbf{0.971} &	1.359 &	1.008 &	1.449 &	\textbf{1.000} &	1.420 &	1.043 &	1.206\\ 
 \hline
Car & \textbf{0.720} &	0.751 &	0.721 &	0.738 &	\textbf{0.729} &	0.763 &	0.767 &	0.806 &	\textbf{0.755} &	0.794 &	0.816 &	0.866\\ 
 \hline
Chair & \textbf{1.322} &	1.387 &	1.335 &	1.367 &	\textbf{1.353} &	1.388 &	1.792 &	1.868 &	\textbf{1.414} &	1.462 &	1.933 &	2.040\\ 
 \hline
Lamp & 1.530 &	2.039 &	\textbf{1.528} &	2.161 &	\textbf{1.571} &	2.028 &	1.657 &	2.163 &	\textbf{1.635} &	2.254 &	1.824 &	2.665\\ 
 \hline
Table & 1.185 &	1.196 &	\textbf{1.181} &	1.196 &	\textbf{1.194} &	1.219 &	1.389 &	1.418 &	\textbf{1.231} &	1.258 &	1.549 &	1.593\\ 
 \hline
\end{tabular}

\end{table*}

\begin{table*}[!htb]
\centering
\caption{\label{tab: inter} Inter-class evaluation of the networks trained on single category and evaluated on another category. Uniformly distributed points are used during training as input point clouds. $AF=8$. \textcolor{black}{The table shows the results of single category training and inter-class evaluation described in section 3.3. The average test loss for each model increases.}}
\renewcommand{\arraystretch}{2}
\begin{tabular}{ |c||c|c|c|c|c|c|c|}
\hline
Training Category & Airplane & Bench & Boat & Car & Chair & Lamp & Table\\ 
\hline
Evaluation Category & \multicolumn{7}{c|}{Test Chamfer Loss  ($\times10^{-3}$)}\\
\hline
\hline
Airplane & \textbackslash & 24.321 & 102.293 & 23.312 & 9.943 & 3.530 & 11.599\\ 
 \hline
Bench & 77.826 & \textbackslash & 35.462 & 44.513 & 1.519 &	6.089 &	1.649\\ 
 \hline
Boat & 3.953 & 11.938 &	\textbackslash &	4.273 &	4.933 &	2.067 &	10.881\\ 
 \hline
Car & 5.152 & 5.406 & 1.543 & \textbackslash & 2.601 & 3.163 &	4.418\\ 
 \hline
Chair & 30.484 & 5.982 & 20.687 & 17.633 & \textbackslash & 11.387 & 4.754\\ 
 \hline
Lamp & 12.609 &	35.19 &	18.575 & 56.709 & 8.846 & \textbackslash & 12.545\\ 
 \hline
Table & 72.059 & 5.909 & 67.493 & 47.852 & 4.883 & 17.199 & \textbackslash\\ 
 \hline
\end{tabular}

\end{table*}

\begin{table*}[!htb]
\centering
\caption{\label{tab: multi} Evaluation of the network trained on a balanced training set involving all seven categories.\textcolor{black}{ The table shows the results of multi-category training and evaluation described in section 3.3. The upsampling condition is the same as Table \ref{tab: inner}. Compared with the results in Table \ref{tab: inter}, this network outperforms all the single-category training networks. Evaluation Chamfer losses on the bench and boat models become less than those in Table \ref{tab: inner}. In addition, the average test loss rises from 1.031 to 1.205 when employing this multi-category training. }}
\renewcommand{\arraystretch}{2}
\begin{tabular}{ |c||c|c|c|c|c|c|c|}
\hline
Evaluation Category & Airplane & Bench & Boat & Car & Chair & Lamp & Table\\ 
\hline
Test Chamfer Loss  ($\times10^{-3}$) & 0.529 & 0.825 & 0.892 & 0.854 & 1.807 & 1.888 & 1.644\\ 
 \hline
\end{tabular}

\end{table*}

\begin{figure*}[!tp]
  \centering
  \includegraphics[trim = 0in 0in 0in 0in, clip, width=\textwidth]{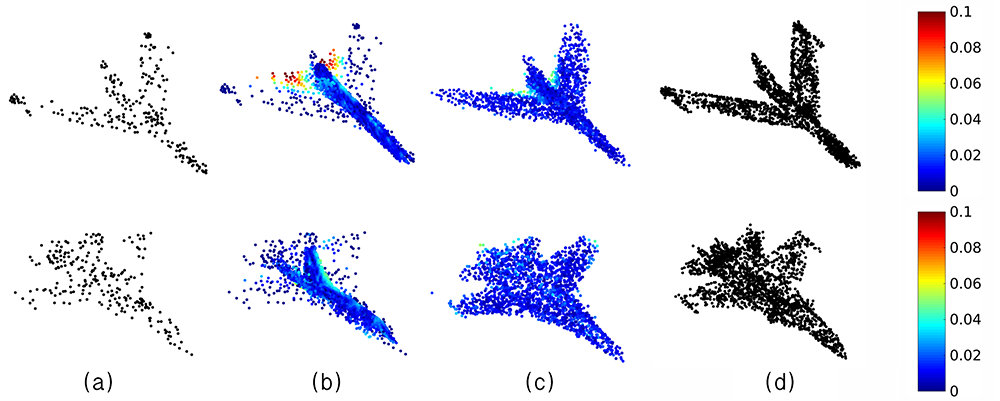}
  \caption{Comparison between our algorithm and EAR method. (a) Input point cloud. (b) Outcome from EAR. (c) Outcome from our algorithm. (d) Ground truth. The color means the distance between a point in the upsampled point cloud and its nearest point in the ground truth point cloud.}
  \label{fig:ear}
\end{figure*}

\begin{figure*}[!tp]
  \centering
  \includegraphics[trim = 0in 0in 0in 0in, clip, width=\textwidth]{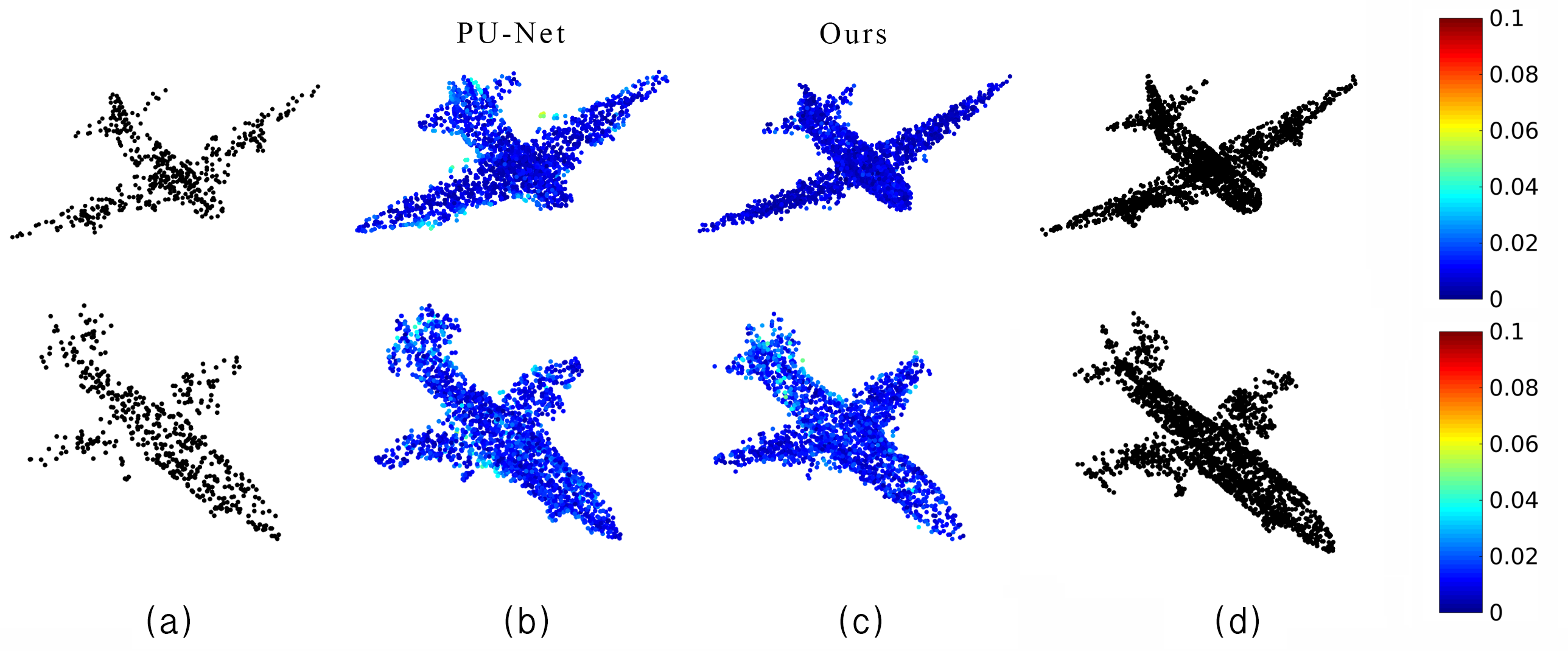}
  \caption{\textcolor{black}{Comparison between our algorithm and PU-Net. (a) Input point cloud. (b) Outcome from PU-Net. The Chamfer Loss for the upper one is 0.000263 and the lower one is 0.000401. (c) Outcome from our algorithm. The Chamfer Loss for the upper one is 0.000150 and the lower one is 0.000406. (d) Ground truth. The color means the distance between a point in the upsampled point cloud and its nearest point in the ground truth point cloud.}}
  \label{fig:pu1}
\end{figure*}

\begin{figure*}[!tp]
  \centering
  \includegraphics[trim = 0in 0in 0in 0in, clip, width=\textwidth]{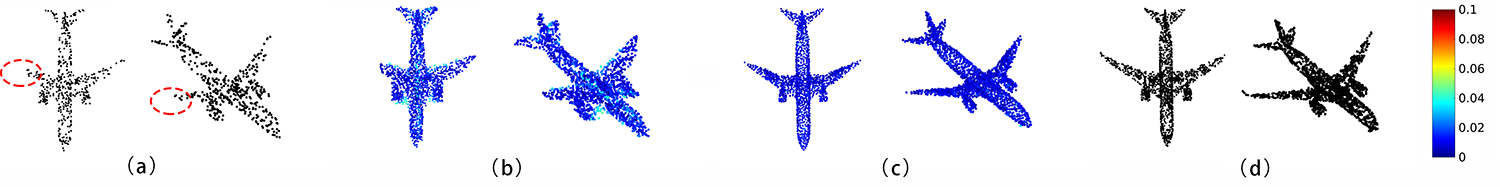}
  \caption{\textcolor{black}{Comparison between our algorithm and PU-Net on point cloud completion. (a) Input point cloud. (b) Outcome from PU-Net. The Chamfer Loss is 0.000779. (c) Outcome from our algorithm. The Chamfer Loss is 0.000212. (d) Ground truth. The color means the distance between a point in the upsampled point cloud and its nearest point in the ground truth point cloud.}}
  \label{fig:pu2}
\end{figure*}

\subsection{Inter-class and multi-class evaluation}
\label{multi}

To further explore the performance of our approach with single-category  and multi-category training, we conduct a case study with $AF=8$, primarily because the largest variance among categories can be observed in Table \ref{tab: inner} under this condition. Additionally, based on the results of Table \ref{tab: inner}, all the input point clouds are sampled from uniform distributions without normal information. Finally, each trained network is evaluated on test cases from the other six categories respectively. The resulting test Chamfer losses are displayed in Table \ref{tab: inter}.

%It is observed that the average loss increases because of the lack of learned experience for other categories. Still, the network trained by chairs achieves $4.877\times10^{-3}$ on average test loss. This might come from the abundant variance in chair models. Bench models are relatively consistent in style and structure with only 5 children types. While chair models, apart from some similar designs, in total include 23 various children types such as arm chairs, folding chairs, recliners or even wheel chairs. Therefore, the network which is trained on chairs and tested on benches, provides less increase percentage in test performance than the network which is trained on benches and tested on chairs. Thus, we can conclude that higher variance in training sets can improve the overall generality of final network model.
%(xxx rewrite this paragraph. the english is poor). 

Expectedly, the average loss increases because the training does not see model features in the other categories used in the evaluation. Nonetheless, evaluation categories that are contextually proximate to the training categories interestingly tend to perform relatively better in the inter-class upsampling (car vs. boat, chair vs. bench, etc.). Surprisingly, the network trained on chairs outperforms the network trained on benches among all the six evaluation categories. In fact, the models in the bench category had only five major subtypes that were largely consistent in style,  whereas the chair database had 23 different subtypes (armchairs, folding chairs, recliners or even wheelchairs). The richness in the chair models results in the least average Chamfer loss ($4.877\times10^{-3}$). Therefore, we can conclude that greater abundance in training sets can improve the generality of network model. 
%xxx rework
% However, higher variance in training sets usually takes more complex training networks and leads to worse performance. To inspect our algorithm on addressing this issue, another network model is trained based on a balanced training set generated from each category and tested on the same test cases as the previous experiments. Referring to table \ref{tab: multi}, comparing with corresponding results in table \ref{tab: inner}, test losses of bench case and boat case even decrease. At the expense that the average test loss rises solely from 1.031 to 1.205, this model is able to upsample point clouds from all 7 categories with relatively high quality and accuracy. More visual results can be found in appendix XXX.

%However, a higher variance in training sets usually requires more complex network architecture or leads to worse performance. To inspect our algorithm on addressing this issue, another network model is trained based on a balanced training set generated from each category. Referring to table \ref{tab: multi}, comparing with corresponding results in table \ref{tab: inner}, test losses of benches and boats even decrease. At the expense that the average test loss rises solely from 1.031 to 1.205, this model is able to upsample point clouds of all 7 categories with relatively high quality and precision. 
%(xxx rewrite this paragraph. the english is poor). 

\begin{figure}[!bp]
  \centering
  \includegraphics[trim = 0in 0in 0in 0in, clip, width=\columnwidth]{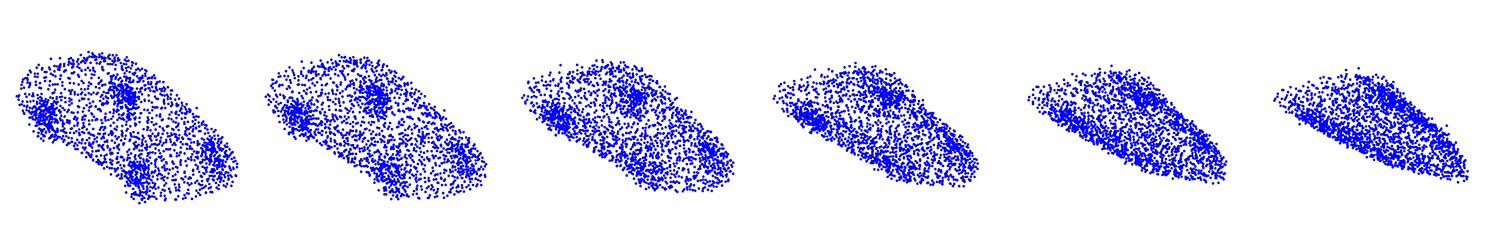}
  \caption{\textcolor{black}{Sample results for shape morphing between the point clouds of a car and a boat.}}
  \label{fig:inter}
\end{figure}

Finally, we trained a single network that uses all seven categories (balanced) for training. Table \ref{tab: multi} shows the performance of this network. Compared with the results in Table \ref{tab: inter}, this network outperforms all the single-category training networks as it learns a richer set of latent features emanating from different categories. Evaluation Chamfer losses on the bench  and boat models become even less than those in Table \ref{tab: inner}. In addition, the average test loss rises merely from 1.031 to 1.205 when employing this multi-category training.
\textcolor{black}{To illustrate the category information learned from multi-category learning, we utilize the feature vectors generated by the this network for shape morphing. A sample result is given in figure 7. We achieved intermediate feature vectors by linearly interpolating feature vectors obtained from the point clouds of a car and a boat. The shape morphing results, shown from left to right in Figure 7, are achieved by varying the weight for the feature vector of the boat from 0 to 1 with a increment of 0.2. It shows that our network learns a unified representation for different categories and is capable of fulfilling a smooth morphing between the global features of two different categories.}

% Though the titantic difference in shapes from

\subsection{Further discussion}\label{further}
% According to the hypothesis we propose in section \ref{inner eval}, a mixture sampling method is designed to provide a potential better input point cloud distribution for upsampling. We introduce a new parameter, combining ratio $\alpha\in[0,1]$, to represent the percentage of points sampled from curvature-based distribution. The other points are sampled from uniform distribution without repetition. As a parametric study, we create 11 group of input point clouds from airplane category with $AF=8$ and an increment of $0.1$ in $\alpha$ each time. The test cases are identical to those we use in table \ref{tab: inner}. Figure XXX summarizes the relationship between $\alpha$ and the corresponding test chamfer loss. As shown, $\alpha=0.1$ or $\alpha=0.2$ both provide better upsampling quality over the pure uniform sampling method ($\alpha=0$). The results exhibit the significance of points near the edges on generalizing the whole shape of the point cloud. Within a certain range of mixture (less than about $20\%$ in this case), the points sampled from curvature-based distribution helps to decrease the chamfer distance in that the features around edges and corners are better represented. But the input point is also becoming less uniform as $\alpha$ get larger. Eventually, the improvement resulted from edge points cannot make up for the increase in chamfer loss caused by too few points on the flat surfaces. We can observe a gradual growth after a critical point ($20\%$) in the curve shown in figure XXX.

\begin{figure}[!tp]
  \centering
  \includegraphics[trim = 0in 0in 0in 0in, clip, width=\columnwidth]{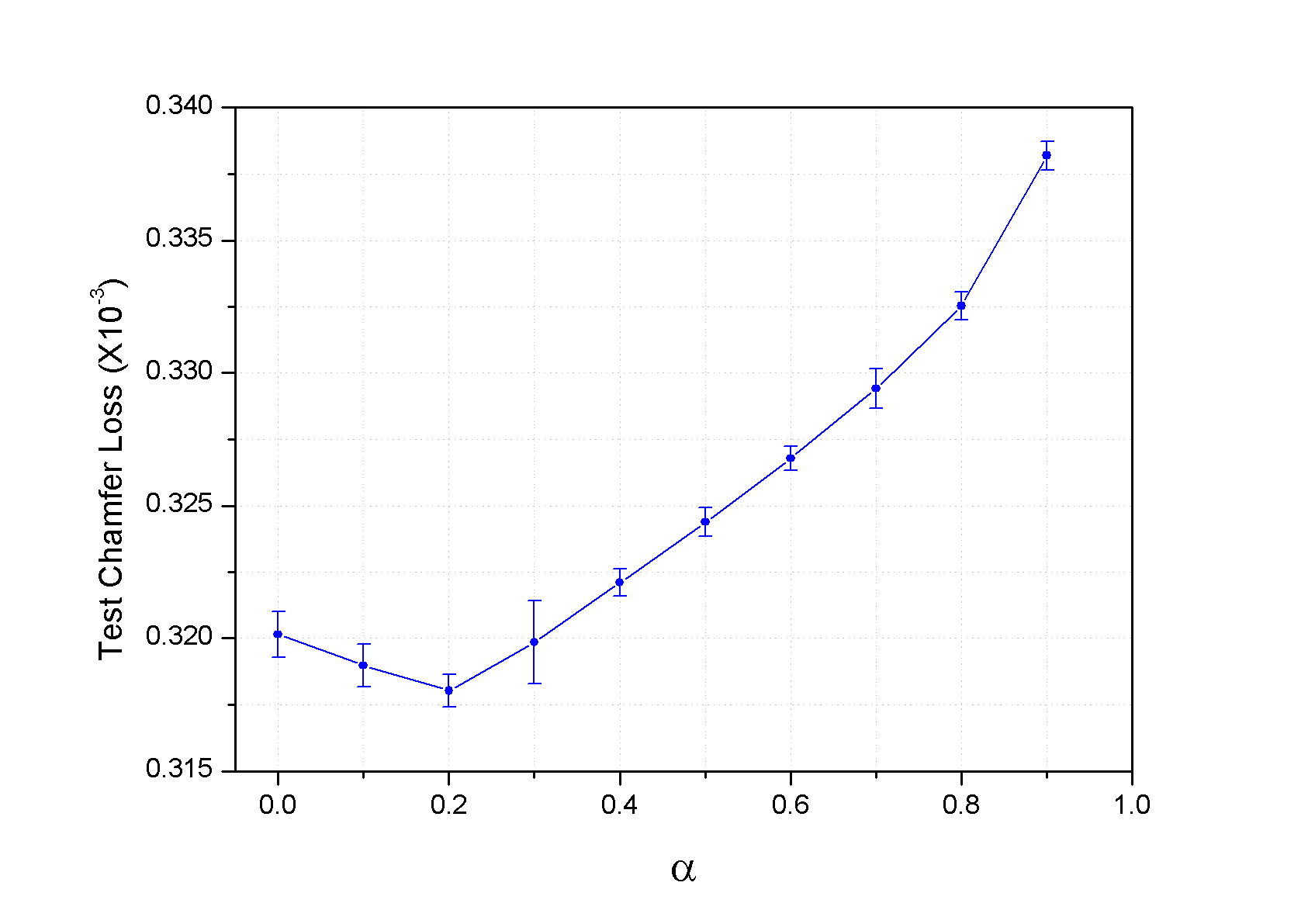}
  \caption{Test Chamfer loss for our trained networks as a function of $\alpha$. \textcolor{black}{ $\alpha\in[0,1]$, that represents the fraction of points subsampled using the curvature-based strategy relative to all subsampled points (the rest being the uniformly subsampled points). As shown, when $\alpha=0.1$ or $\alpha=0.2$, both provide higher upsampling quality over the uniformly sampled points ($\alpha=0$). As seen, with approximately $20\%$ of the subsampled points coming from the curvature-based approach, the average Chamfer loss is minimized.}} %(xxx change combine ratio to $\alpha$ in the x axis)}
  \label{fig:combine}
\end{figure}

Finally, we test a hybrid subsampling method designed to provide a potentially superior input point cloud for upsampling. For this, we introduce a new parameter, $\alpha\in[0,1]$, that represents the fraction of points subsampled using the curvature-based strategy relative to all subsampled points (the rest being the uniformly subsampled points). We create 11 groups of input point clouds from the airplane category with $AF=8$ and an increment of $0.1$ in $\alpha$ each time. The test cases are identical to those we use in Table \ref{tab: inner}. Figure \ref{fig:combine} shows the relationship between $\alpha$ and the corresponding test Chamfer loss. As shown, when $\alpha=0.1$ or $\alpha=0.2$, both provide higher upsampling quality over the uniformly sampled points ($\alpha=0$). As seen, with approximately $20\%$ of the subsampled points coming from the curvature-based approach, the average Chamfer loss is minimized. This implies that there exists a trade-off between using purely curvature-based versus purely uniform subsampled points, with the hybrid approach providing a more desirable outcome. Note that as $\alpha$ increases, the input point cloud becomes much less uniform, and eventually, the improvement resulting from the addition of edge and feature-rich regions cannot make up for the lack of points in the flat (low curvature) regions. The accuracy and coverage values reported in \ref{app: acc_cov} further verify this inference. 

% As a comparison with the model we obtained from the best combining ratio, we utilize EAR \cite{XXX}, a state-of-the-art method for optimization-based point cloud upsampling method, to complete the same task. Figure XXX illustrates that our best model provides more precise upsampling results over EAR method. 

As a comparison with the model we obtained using this hybrid subsampling model, we utilize EAR \cite{Huang:2013:EPS:2421636.2421645}, a state-of-the-art method for optimization-based point cloud upsampling, toward the same task. Figure \ref{fig:ear} shows that our algorithm provides more precise upsampling results over the EAR method. Since the number of points is small, the EAR method only finds only some of the major edges and uses most of the points around those regions. 

\textcolor{black}{We also compare our method with a patch-level learning method, PU-Net \cite{DBLP:journals/corr/abs-1801-06761}. In Figure \ref{fig:pu1}, we show the reconstruction results based on our method and PU-Net. Judging from the Chamfer loss, our method and PU-net are compatible. However, Figure \ref{fig:pu2} demonstrates that our method evidently outperforms PU-Net on point cloud shape completion. In this experiment, we remove half of the right wing in the given input airplane point cloud. PU-Net is only able to add more points around the existing points without any shape completion. While, our method manages to upsample the incomplete input point cloud with smooth completion, because object-level and category-based learning enables the network to sense the underlying object and generalize the missing features based on learned global features frequently appearing in this category. This advantage of shape completion can be favorable in practical deployment since the scanning point clouds usually cannot cover all the critical features and details.}

%%%%%%%%%%%%%%%%%%%%%%%%%%%%%%%%%%%%%%%%%%%%%%%%%%%%%%%%%%%%%%%%%%%%%%%%
\section{Conclusions}

This work presents a deep learning algorithm aiming at upsampling a sparse point cloud with a prescribed amplification factor determined by the user. Instead of using  human defined priors or heuristics, we exploit the deep networks' ability to extract the latent features for upsampling for various object categories. Then these latent features assist the point cloud upsampling so that a common global feature set learned from a single or a multitude of object categories can be naturally utilized.

We successively explore the effect of two different distributions for input point cloud sampling. Based on the outcomes, a further parametric study on the hybrid sampling ratio for points produced by these two sampling strategies is conducted to identify the benefit of using feature-sensitive points for upsampling. As this ratio increases, points sampled on the high-curvature regions of an object is capable of better capturing the critical feature-rich regions. Meanwhile, a reduction of the point density in areas far from the feature regions causes a scarcity of points in these areas during upsampling, which expectedly decrease the accuracy in the current loss function. Nonetheless, in real applications, fewer points in the flat, featureless regions may not be a  severe issue, since by and large most current surface reconstruction methods assume smooth surfaces between the sampled points. As such, our future work will explore alternative metrics to assess the resulting upsampled point cloud so as to enable non-uniform but high quality upsampled models. 
%%%%%%%%%%%%%%%%%%%%%%%%%%%%%%%%%%%%%%%%%%%%%%%%%%%%%%%%%%%%%%%%%%%%%%%%
\section{Limitations and Future Work}
% Our current model is trained on point clouds sampled from PointNet. As we inspect the meshes utilized in our training, we find out that part of the meshes have flaws like open meshes or double-sided surface. These flaws lead to some incorrect normal calculation, which is a potential reason why normal information doesn't enhance the upsampling quality. An alternative way is to obtain the normal vectors only depending on the vertex coordinates. We believe it will bypass the defect in mesh surfaces and provide more precise normal vectors so that we can extensively verify its influence.  

% Another limitation of our study is that our deep network is proficient at reconstruct the global features that are frequently occurred in the category. For those features that rarely exist or never occur in the training sets, the upsampling quality is not as good as the former ones. To address this problem, we can investigate the possibility of replenishing latent features learned from patch scale.  

Our current model is trained on point clouds sampled from ShapeNetCore. An inspection of the available models reveals that parts of the meshes have geometric shortcomings such as open disconnected polygons or double-sided surfaces. These flaws lead to incorrect normal calculations, which is a potential reason why the normal information did not contribute positively to the upsampling quality. An alternative way is to compute normal vectors using the available vertex coordinates through local patch fitting, which is to be explored in the future.

Another characteristic of our approach is that while it performs well at reconstructing the global features that are frequently occurred in the category,  the upsampling quality diminishes for features that are rarely encountered during training. To address this problem, we plan to investigate the possibility of replenishing our network with latent features learned at the local, patch scale.

Finally, our tests thus far have concentrated on models of man-made, engineered objects. Whether the same approach generalizes well for object categories consisting of natural or biological models is the subject of future work. 
%%%%%%%%%%%%%%%%%%%%%%%%%%%%%%%%%%%%%%%%%%%%%%%%%%%%%%%%%%%%%%%%%%%%%%%%
%\section{Acknowledgement}

%%%%%%%%%%%%%%%%%%%%%%%%%%%%%%%%%%%%%%%%%%%%%%%%%%%%%%%%%%%%%%%%%%%%%%%

\label{}
% Commenting out the following part is to make loading faster

\newpage
\appendix

\section{Accuracy and Coverage}
\label{app: acc_cov}
We use the same point cloud upsampling metrics introduced in  \cite{DBLP:journals/corr/AchlioptasDMG17} to illustrate the performance of our algorithm. \textcolor{black}{That is (a) Accuracy: which is the fraction of the predicted points that are within a given radius  ($\rho$) from any point in the ground truth point cloud and (b) Coverage: which is the fraction of the ground-truth points that are within $\rho$ from any predicted point.} Referring to Table \ref{tab: acc} and Table \ref{tab: coverage}, as a supplement to Table \ref{tab: inner}, the accuracy and coverage values are reported with $\rho=0.03$.

We explain the principle behind Figure \ref{fig:combine} in section \ref{multi}. Here, the accuracy and coverage values are reported in Table \ref{tab: combine}. Since accuracy and coverage are both  high (98.5\%) and close in both cases when $\rho=0.03$, we use $\rho=0.015$ in this case.

\begin{table*}[bp]
\centering
\caption{\label{tab: combine}Results of accuracy and coverage for networks as a function of $\alpha$.}
\renewcommand{\arraystretch}{2}
\begin{tabular}{|c||c|c|c|c|c|c|c|c|c|c|c|}
\hline
$\alpha$ & 0.0   & 0.1   & 0.2   & 0.3   & 0.4   & 0.5   & 0.6   & 0.7   & 0.8   & 0.9   & 1.0   \\ \hline
Accuracy (\%)  & 80.25 & 80.34 & 80.34 & 80.24 & 80.21 & 79.96 & 79.78 & 79.69 & 79.65 & 79.55 & 79.12 \\ \hline
Coverage (\%)  & 83.05 & 82.92 & 83.03 & 82.81 & 83.02 & 82.89 & 82.53 & 82.55 & 82.45 & 82.35 & 81.55 \\ \hline
\end{tabular}
\end{table*}

\begin{table*}[h]
\centering
\caption{\label{tab: acc} Accuracy of networks trained on seven categories respectively under twelve conditions.}
\renewcommand{\arraystretch}{2}
\begin{tabular}{ |c||c|c|c|c|c|c|c|c|c|c|c|c|}
\hline
\multirow{3}{*}{} AF & \multicolumn{4}{c|}{2} & \multicolumn{4}{c|}{4} & \multicolumn{4}{c|}{8}\\ 
\cline{1-13}
                    Sample & \multicolumn{2}{c|}{U} & \multicolumn{2}{c|}{CB} & \multicolumn{2}{c|}{U} & \multicolumn{2}{c|}{CB} & \multicolumn{2}{c|}{U} & \multicolumn{2}{c|}{CB}\\
\cline{1-13}
                    Normal & No & Yes & No & Yes & No & Yes & No & Yes & No & Yes & No & Yes\\
\hline
\hline
Category & \multicolumn{12}{c|}{Accuracy ($\%$)}\\
\hline
Airplane & 98.94 & 98.85 & 98.94 & 98.69 & 98.83 & 98.6 & 98.62 & 98.26 & 98.48 & 98.19 & 98.6 & 97.45\\
 \hline
Bench & 92.53 &	89.36 &	92.8 & 66.61 & 92.57 & 89.78 & 91.98 & 88.71 & 92.1 & 89.03 & 91.47 & 87.73\\ 
 \hline
Boat & 90.1 & 82.3 & 89.79 & 81.71 & 89.62 & 82.76 & 89.09 & 81.47 & 89.31 & 82.16 & 87.35 & 79.98\\ 
 \hline
Car & 91.94 & 91.45 & 91.91 & 91.77 & 91.83 & 91.24 & 91.18 & 90.45 & 91.39 & 90.56 & 90.53 & 89.64\\ 
 \hline
Chair & 80.17 &	79.11 &	80.02 &	79.63 &	79.7 &	78.96 &	74.56 &	73.6 & 78.98 & 78.33 & 72.91 & 70.94\\ 
 \hline
Lamp & 77.41 & 71.15 & 77.47 & 70.33 & 76.65 & 70.96 & 75.95 & 69.79 & 75.63 & 68.06 & 73.88 & 64.02\\ 
 \hline
Table & 87.15 &	86.81 &	86.87 &	86.58 &	86.79 &	86.35 &	85.13 &	84.63 &	86.22 &	85.87 &	83.79 &	83.3\\ 
 \hline
\end{tabular}
\end{table*}

\begin{table*}[h]
\centering
\caption{\label{tab: coverage} Coverage of networks trained on seven categories respectively under twelve conditions.}
\renewcommand{\arraystretch}{2}
\begin{tabular}{ |c||c|c|c|c|c|c|c|c|c|c|c|c|}
\hline
\multirow{3}{*}{} AF & \multicolumn{4}{c|}{2} & \multicolumn{4}{c|}{4} & \multicolumn{4}{c|}{8}\\ 
\cline{1-13}
                    Sample & \multicolumn{2}{c|}{U} & \multicolumn{2}{c|}{CB} & \multicolumn{2}{c|}{U} & \multicolumn{2}{c|}{CB} & \multicolumn{2}{c|}{U} & \multicolumn{2}{c|}{CB}\\
\cline{1-13}
                    Normal & No & Yes & No & Yes & No & Yes & No & Yes & No & Yes & No & Yes\\
\hline
Category & \multicolumn{12}{c|}{Coverage ($\%$)}\\
\hline
\hline
Airplane & 99.05 & 98.81 & 98.93 & 98.63 & 98.9 & 98.58 & 98.54 & 98.06 & 98.43 & 98 & 98.58 & 96.74\\
 \hline
Bench & 90.3 &	84.99 &	90.96 & 73.82 & 90.26 & 85.74 & 89.31 & 83.37 & 89.23 & 82.81 & 88.09 & 82.1\\ 
 \hline
Boat & 91.74 & 84.92 & 91.89 & 84.19 & 91.19 & 84.24 & 90.6 & 82.65 & 90.74 & 83.58 & 89.27 & 81.78\\ 
 \hline
Car & 94.64 & 93.83 & 94.38 & 94.16 & 94.43 & 93.53 & 93.46 & 92.68 & 93.94 & 92.71 & 92.59 & 91.42\\ 
 \hline
Chair & 81.26 &	79.71 &	80.51 &	80.05 & 80.48 &	80.18 &	76.76 &	75.52 & 79.25 & 78.5 & 75.02 & 73.48\\ 
 \hline
Lamp & 84.44 & 76.97 & 84.72 & 76.84 & 84.45 & 78.28 & 83.2 & 76.8 & 83.78 & 75.77 & 81.95 & 71.94\\ 
 \hline
Table & 81.4 &	81.15 &	81.39 &	81.04 &	81.09 &	80.72 &	79.16 &	78.41 &	80.64 &	79.97 &	77.17 &	77.07\\ 
 \hline
\end{tabular}
\end{table*}

\section{Benches vs Chairs}
\label{app: ben_cha}

\begin{figure*}[hp]
  \centering
  \includegraphics[trim = 0in 0in 0in 0in, clip, width=\textwidth]{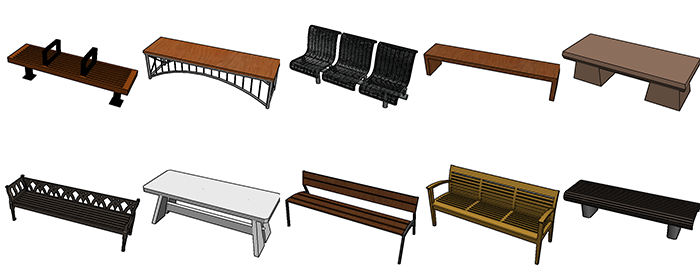}
  \caption{Ten randomly selected benches in the dataset.}
  \label{fig:bench}
\end{figure*}

\begin{figure*}[hp]
  \centering
  \includegraphics[trim = 0in 0in 0in 0in, clip, width=\textwidth]{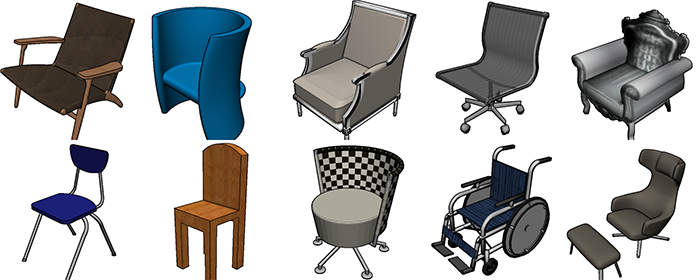}
  \caption{Ten randomly selected chairs in the dataset.}
  \label{fig:chair}
\end{figure*}

To demonstrate the variance in the chair and bench dataset, we randomly select 10 models in each category. The textured models are shown in Figure \ref{fig:bench} and Figure \ref{fig:chair}. Larger shape variation in the chair models can be observed.

\section{Upsampling Results}
\label{more results}
\begin{figure*}[th]
  \centering
  \includegraphics[trim = 0in 0in 0in 0in, clip, width=\textwidth]{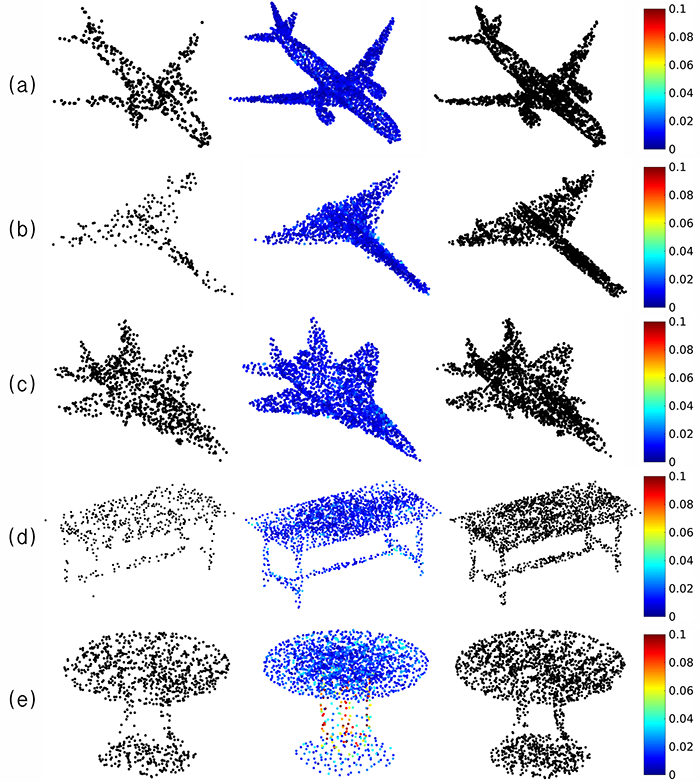}
  \caption{Sample results (Part I). (a) $AF=4$, U, ST. (b) $AF=8$, U, ST. (c) $AF=2$, U, ST. (d) $AF=4$, U, ST. (e) $AF=2$, CB, ST.}
  \label{fig:s1}
\end{figure*}

\begin{figure*}[th]
  \centering
  \includegraphics[trim = 0in 0in 0in 0in, clip, width=\textwidth]{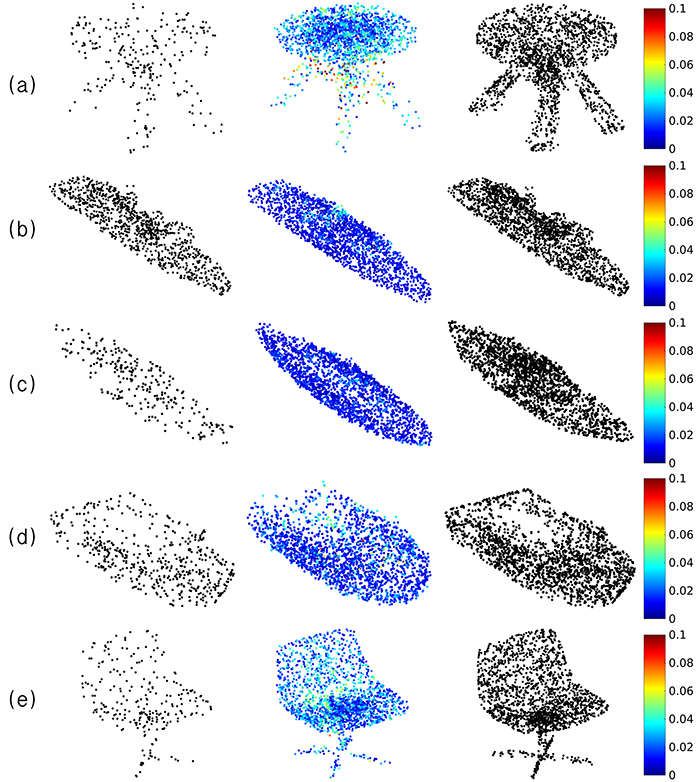}
  \caption{Sample results (Part II). (a) $AF=8$, U, ST. (b) $AF=2$, CB, ST. (c) $AF=8$, U, ST. (d) $AF=4$, U, ST. (e) $AF=8$, U, ST.}
  \label{fig:s2}
\end{figure*}

\begin{figure*}[th]
  \centering
  \includegraphics[trim = 0in 0in 0in 0in, clip, width=\textwidth]{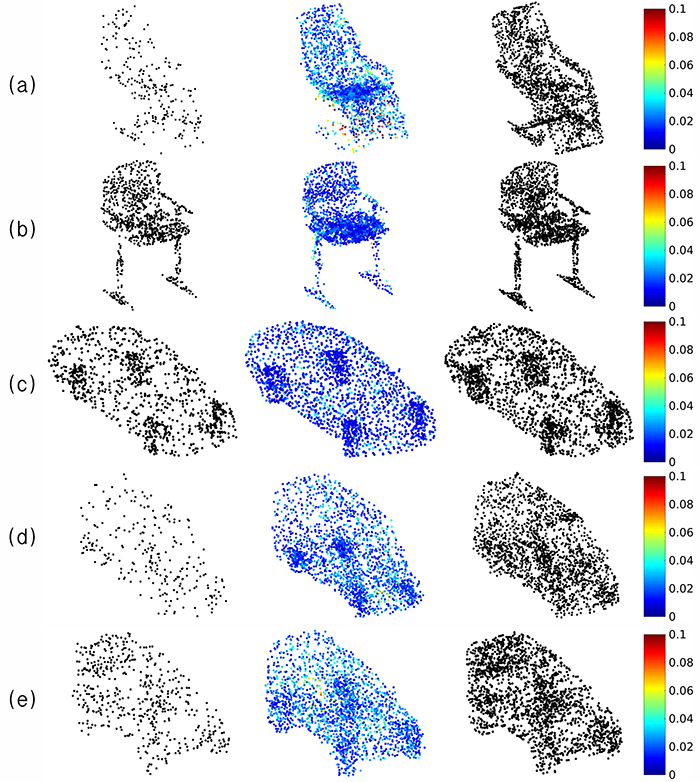}
  \caption{Sample results (Part III). (a) $AF=8$, CB, ST. (b) $AF=2$, U, ST. (c) $AF=2$, U, ST. (d) $AF=8$, U, ST. (e) $AF=4$, U, ST.}
  \label{fig:s3}
\end{figure*}

\begin{figure*}[th]
  \centering
  \includegraphics[trim = 0in 0in 0in 0in, clip, width=\textwidth]{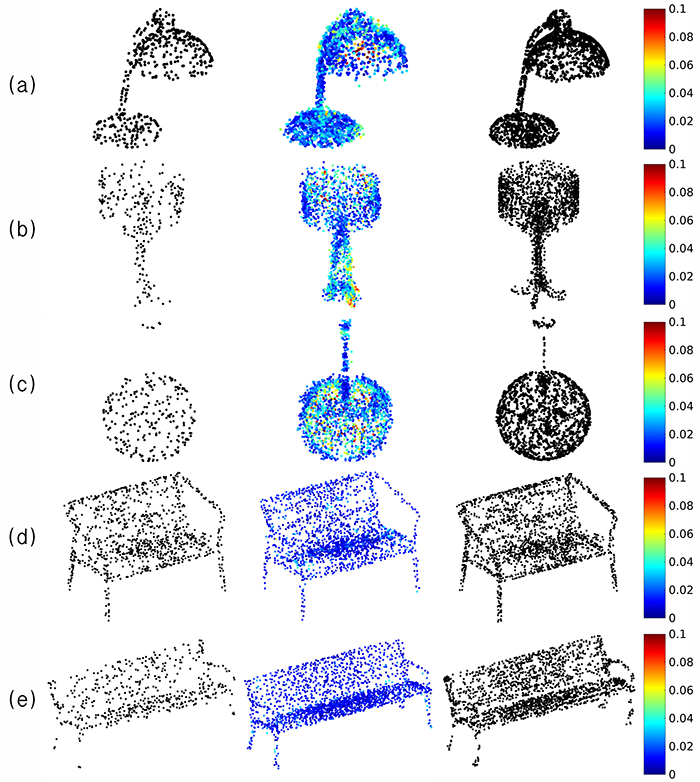}
  \caption{Sample results (Part IV). (a) $AF=4$, U, ST. (b) $AF=8$, U, ST. (c) $AF=8$, U, ST. (d) $AF=2$, CB, ST. (e) $AF=4$, U, ST.}
  \label{fig:s4}
\end{figure*}

\begin{figure*}[th]
  \centering
  \includegraphics[trim = 0in 0in 0in 0in, clip, width=\textwidth]{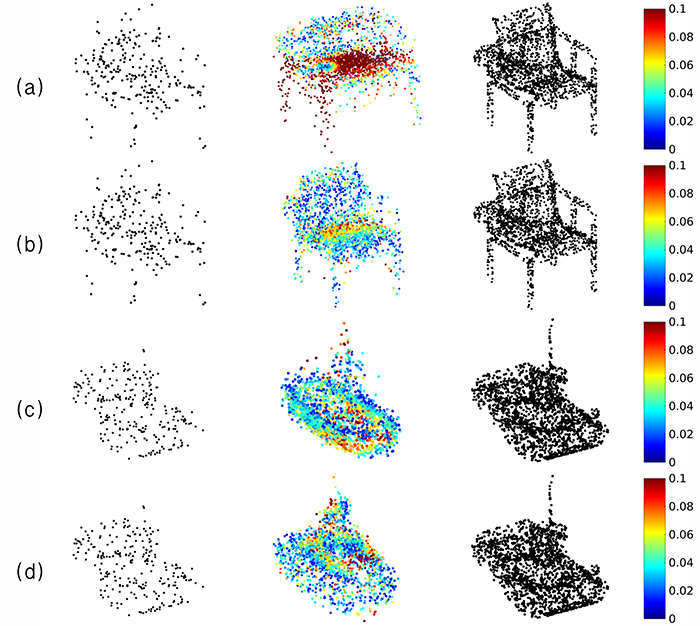}
  \caption{Sample results (Part V). (a) $AF=8$, U, ST. (b) $AF=8$, U, MT. (c) $AF=8$, U, ST. (d) $AF=8$, U, MT.}
  \label{fig:s5}
\end{figure*}

Further upsampling results are shown in Figure \ref{fig:s1}, \ref{fig:s2}, \ref{fig:s3}, \ref{fig:s4} and \ref{fig:s5}. In all cases, the left column is the input sparse point cloud, the middle column in our output, and the right column is the ground truth. 

Details for each case are listed in the figure caption.
(U: Uniform sampling; CB; Curvature-based sampling; ST: Single-category training; MT: Multi-category training. In all cases, we don't use normal information.)
% Figure \ref{fig:s1}:
% (a) $AF=4$, U, ST. (b) $AF=8$, U, ST. (c) $AF=2$, U, ST. (d) $AF=4$, U, ST. (e) $AF=2$, CB, ST.
% Figure \ref{fig:s2}:
% (a) $AF=8$, U, ST. (b) $AF=2$, CB, ST. (c) $AF=8$, U, ST. (d) $AF=4$, U, ST. (e) $AF=8$, U, ST.
% Figure \ref{fig:s3}:
% (a) $AF=8$, CB, ST. (b) $AF=2$, U, ST. (c) $AF=2$, U, ST. (d) $AF=8$, U, ST. (e) $AF=4$, U, ST.
% Figure \ref{fig:s4}:
% (a) $AF=4$, U, ST. (b) $AF=8$, U, ST. (c) $AF=8$, U, ST. (d) $AF=2$, CB, ST. (e) $AF=4$, U, ST.
% Figure \ref{fig:s5}:
% (a) $AF=8$, U, ST. (b) $AF=8$, U, MT. (c) $AF=8$, U, ST. (d) $AF=8$, U, MT.

\end{document}